# Machine Learning and the Future of Realism

Giles Hooker, Cliff Hooker[1]

**Contents**

1. *Machine learning, its rise and nature*

The preceding three decades have seen the emergence, rise, and proliferation of machine learning (ML). From half-recognised beginnings in perceptrons, neural nets, and decision trees, algorithms that extract correlations (that is, patterns) from a set of data points have broken free from their origin in computational cognition to embrace all forms of problem solving, from voice recognition to medical diagnosis to automated scientific research and driverless cars, and it is now widely opined that the real industrial revolution lies less in mobile phone and similar than in the maturation and universal application of ML. Among the consequences just might be the triumph of anti-realism over realism.

A venerable and widely familiar example of an ML tool is a so-called neural net. The neural net learning algorithm in simple form involves procedures for representing inputs (data points) as loads on net-input layer nodes. These loads are then redistributed across nodes of the next layer according to a weighting formula on node connections. This redistributing operation is repeated on successive internal layers until the last (output) layer is reached. Various node connection weightings are trialled on a training data set until the net yields the appropriate outputs (until it has learned its task) and then it can be applied predictively to further, related data sets. In this fashion nets have learned many tasks easily that continue to be difficult and sometimes impossible for humans. However, only the initial (input or setup) and last (output) layers are cognitively interpretable; in general, the remainder cannot be assigned any cognitively meaningful state. Just this is the secret of their success. The net learning algorithm specifies the states of internal nodes and the functions that change them numerically, not logically, so that net processing is sub-categorical with respect to the input and output psychological categories. The output content is not arrived at from the input by any process of concept-preserving logical analysis. Rather, just this shift greatly increases accessible net states and state change functions. Using nets involves rejecting the long and widely-held assumption of adhering to agency categories when constructing psychological theory.[1]

---

[1] Giles Hooker is Associate Professor of Biological Statistics and Computational Biology at Cornell University. Cliff Hooker is Professor Emeritus of Philosophy at the University of Newcastle (Australia).

[1] For an introduction to machine learning see, e.g., Wikipedia, 2016.. Cf. Ray 2015. For the sub-categorical nature of the neural net algorithm see Smolensky, 1988. For more on neural nets see Churchland 1989, Hooker 1993.

Neural nets are not the only ML algorithms—the overview Wikipedia entry lists 50+ forms (see n. 1). Other ML algorithms are based on quite different methods that do not have the same (mostly false) biological motivation. But they do highlight one challenging feature widely (but not universally) shared among such algorithms: the absence of interpretability for their internal states. This absence can also extend to their outputs as well, if, for example, the output, and the internal states, are expressed as an infinite dimensional vector. Nonetheless, when the data sets concern the behaviour of entities under study, the algorithms deliver, via their estimated correlation relationships, predictions of entity behaviour as accurate as the data set permits and, as the data set increases, accuracy as high as any model could achieve. Note again that a learning algorithm will produce "naked predictions," that is, numbers stripped of any ontological interpretation: it is not that they present a different ontological interpretation, or that one will appear once the data are "cleaned up"; they offer none at all. Their predictive success combined with their a-ontolological status stands at the heart of the challenge they pose to realism.

2. *Machine Learning supports the centrality of a-model naked prediction*

The usual context in which prediction is discussed in philosophy of science is one where predictions are derived by deduction from initial data conjoined to some theoretical model that is interpreted (realized) in terms of that theory's ontology, whether that be atomic or field, dynamical or functional, etc. When such a model is not available, neither is prediction. But there are myriad cases where this requirement is not satisfied, where we can determine that the system has changed but there is no satisfactory interpreted model of how that happens. Examples include processes too complex to follow, as occurs in complex systems where dynamical form may transform, and critical and chaotic regimes arise, and/or data too vast for humans to comprehend, as occurs with increasing frequency (in big business, in bio-molecular science, etc.). ML offers naked prediction in every case, whether or not those predictions also follow from an interpreted model, and is often our only way to advance research and understanding.

Given this, naked prediction seems more central to science than modelling does. Why not then accept that science is based solely on naked prediction, ignore the call to understand the theoretical world beyond this as a siren call, and argue that the diverse goals of science (prediction, explanation, unification, etc.) can all be parsed in terms of naked predictive performance alone? Note that this is not a proposal to remove this or that class of models, but rather the claim that it is not necessary to use theoretical models at all, indeed not necessary to use theory at all (Breiman 2001). And with that goes any claim that science guides choice of ontology. The centrality of a-model naked prediction is by default an anti-realist position.

The only commitments to human-scale empirical elements (input and outcome data, measuring instruments, etc.), are those required by a recurring phenomenological empiricism. ML still requires a human to identify a prediction task, choose a competent ML algorithm, specify relevant inputs and outputs, and procure adequate training and an experimental data set. These are not at present interpretation-free activities, which raises the prospect that ML illegitimately smuggles in interpretive models.[2] Some entertain finessing the problem by use of all possible

---

[2] An important (to humans) aspect of this is the issue of how to deal with ethical and social biases that are exhibited in the data sets we give it, e.g., prejudicial attitudes toward, and institutional barriers to, opportunities for coloured

data of all possible kinds to produce a universal correlation web. We will return to this issue in §5 below. Meanwhile, the point remains that ML does allow us to avoid an interpretable characterization of the relationship between these inputs and outputs once they are defined.

3. *History of this cleavage*

Science has been constantly faced with this issue. It is, for example, the motivating division behind those who, a century and more ago, pushed for a science of psychology founded on revealing the inner structure and contents of minds (e.g., Wundt) and those who urged confinement to just extracting patterns within stimuli and responses (e.g., Skinner). But back then it was a matter of high philosophical differences that carried the debate. Now the development of ML forces this issue in an immediate, unavoidable way. The wheel has also turned full circle: Skinnerian empiricism was replaced by computational cognitivism's quasi-internalist modelling, only to be challenged in turn by ML's externalist agnosticism. More broadly, this methodological difference is the deepest expression of the realist/anti-realist cleavage that runs through the philosophy of science.[3]

4. *The central question and its answer*

Is science satisfactorily characterizable in terms of prediction alone ("naked prediction") or is there something more that is essential to science? A second question follows: By what criteria ought the preceding question be decided? A popular but reasonable response to this latter question is that any additional end proposed for science ought to "pay its way," i.e., to add something that is valuable independently of the value that naked prediction brings. Naked prediction serves the inherent epistemic value of naked pattern knowledge, and with that naked explanation, and also the pragmatic value of control, we can manipulate the world (including ourselves) to achieve various pragmatically valuable ends from engaging in research to building bridges and curing maladies.

What then of any ends of science beyond what prediction supports? The most compelling answer, we suggest, is the provision of an intelligible conception of deep or underlying being (ontology). What sort of being or beings is the cosmos? How does it work? How are living beings constituted as living? What are their ends (if any) inherent in their being? These and related questions are all central to our personal and collective drive for enlightenment. If anything beyond manipulative or naked pattern knowledge has inherent value, it is ontological enlightenment. This notion is tied to theoretically interpretable ontology, and hence to realist interpretable modelling.

5. *Supports, persuasive and not, for interpretable modelling*

---

people in the US. Do we allow ML to perpetuate these biases? See the recent workshop on Machine Learning for Social Good (Faghmous *et al.* 2016). It is possible to design ML to avoid these, but only with the understanding that they need to be avoided.

[3] See, e.g., van Fraassen versus realists in Churchland and Hooker 1985, and Hooker's recent review in Hooker 2011, §6.2.5.

*Simplicity*. Might simplicity considerations support interpretable modelling? ML is vulnerable to an Occam's razor-type argument in that its models are more complex than needed if an interpretable model will do the job. But the prediction-alone approach equally avoids commitment to any ontological categories beyond the empirical phenomena of experimentation, etc. noted above and in any case these categories are shared with interpretable modelling. And with that, the prediction-alone approach also avoids evaluating theory coherency, historical sequences of theories that might support realism, and much more (cf. n. 3). Moreover, as input-output data increases, the ML output converges on the predictions of a true model, any differences of predictive accuracy go away and any differences in output simplicity go with them. That is, arguments about simplicity (or, for instance, Popperian falsifiability criteria) are tacitly predicated. Our conclusion is that simplicity is too complex, so it does not produce a useful test.

*Risk*. ML similarly cannot convincingly claim that it takes on less risk than does interpretable modelling (or vice versa). While ML's prediction-only approach avoids the ontological risks of error noted above, it still risks using an inadequate method to resolve small differences between predictions by not including relevant input sets and/or by not having sufficient internal discrimination (e.g., a neural net with too few nodes per layer) or by having insufficient or noisy data. (This is the problem faced in §2.) However, lacking any guidance from what is there (being ontologically non-commital), the only option available is to include all possible input data to an infinitely internally diverse algorithm and to trust it will remove the risks. These requirements are certainly in principle beyond practical realization, especially for finite beings that commence in ignorance and remain fallible, but is such an algorithm nonetheless in principle able to do the job if realized? We suppose not, but leave the question open.[4] Either way, our conclusion is that risk is also too complex, so it does not produce a useful test.

*Efficiency*. Interpretable models provide important categories of information that ML cannot in principle provide, and interpretable models provide it in immediately intelligible and usable forms. In general, interpretable models tell us about the states and processes that studied systems exhibit, while ML can only characterize input-output relationships. I can transfer the force that holds our solar system together, found out from its correlational data, to any other solar system without alteration, and so start working out its physics immediately. However, it will be replied, whenever practical use is to be made of this information, whether for astro- or astrodome physics, it always suffices directly to use the relevant input-output relationships. If so, the intermediary details are merely one way to arrive at the outputs but are no more essential to doing so than are the particular internal details of a competent ML algorithm. Thus far our preference for interpretable details is just that: an affectation or cultural habit that does no independent work.

There is this further consideration. An interpretable model comes with a framework of dynamical modalities: possibility, actuality, necessity, including super-possibilities and super-necessities for how they may change their specific versions of these, for example under bifurcations. What a fundamental theory of physics primarily tells us is what the dynamical possibilities and necessities are in some domain, and the rest is input-output relations (called flows or vector fields). ML algorithms, being confined to naked data and their relationships, cannot in principle provide the modal information.[5] But again comes the response that, while modality may make it easier for us to make predictions, it is also unnecessary for that. Like our interest in states and processes, our interest in modality is just that: a pragmatic preference deriving from ease of human use when that was an important consideration, perhaps, but not a distinction that does any independent work. But before acquiescing, note one potential sticking point: some complex dynamical systems show bifurcation, a sharp change in dynamical form itself that amounts to change in what is dynamically possible and necessary, and that raises the question of whether ML can adequately characterize these changes (with whatever internal discriminators it uses). There may be reason to think not, and if so we have one ineliminable functional argument for retaining interpretable models in addition to their inherent value for us.[6]

---

[4] Cherniak 1986 is a standing rebuke to anyone tempted to think that discriminating what is "in principle" possible (impossible, necessary, etc.) is straightforward.

*Unification*. The value of ontological understanding is supported, we contend, by the partial but deep successes of a closely-related scientific end: theoretical unification. A unification requires provision of a single unifying ontology, thus driving ontological understanding deeper through transformation of the possibility-necessity framework. The unification of electromagnetic and mechanical dynamics to form relativity theory provides a deep revelation of our spatiotemporal being in the world. Again the currently-developing unification of biological chemistry and quantum macromolecular theory will provide a deep understanding of living beings. And so on. However, following the preceding indecisiveness, we must ask whether machine learning can again provide all the relevant input-output relationships. If machine learning is presumed to have no limitations to its capacity to form these relationships, it will develop input-output representations of systems that are pre-unification and systems that are post-unification, but of course as naked inputs and outputs, not characterized in interpretable ontological terms. Again, there is nothing further needed for the use of the predictions. (There remains their inherent value for us.) However, most or all of these inter-theory relations are specified in terms of asymptotic relations ($1/c \to 0$, $h \to 0$, etc.) and these also involve indefinitely fine discriminations, so the preceding doubt (n. 6) remains.

6. *Realism in a partially accessible world*.

We live in a world that quickly becomes epistemically inaccessible to us beyond the simplest systems. We may know bare Newton's equations, but when joined with specific interactions and constraints for actual use, the equations quickly become insoluble in algebraic functions beyond their simpler forms. This bars our way to interpretable knowledge of them. But since we know the equations we can in principle acquire a lesser but still interpretable knowledge of them by numerically approximating their states in sequence beginning from initial inputs. Since we are not in general able to infer parametric models from these data, or even much (though not no) knowledge of possibility and necessity, this is not very different from pure input-output knowledge. (The exception is that the basic components [this lever, those atoms, etc.] will be interpretively known throughout, so long as they are not dissipated by the dynamics.) But this is not the only part-way house of knowledge that can arise. Suppose, for example, that Newton could not deduce the gravitational principle that held the solar system together, but had access to ML to apply to astronomical data. Suppose that he could isolate the samenesses and differences in the internal ML states corresponding to a variety of planets and moons. What could he learn? The inverse square law? No. This is not because the states are naked (we want the spatial pattern, not its ontology), but because what could easily, and would likely, appear is a very complex set of numerical interrelationships taking perhaps hundreds of pages to print out, and making for a wholly uninterpretable, hence inaccessible, representation of gravity. But while we cannot make use of it, the ML device can.

---

[5] This is essentially Seller's critique of anti-realist empiricism (see, e.g., Sellars 1965).

[6] One argument is this: machine learning by definition uses computational algorithms, but all computer programs are finite and so must generate rounding-off errors. Thus dynamical processes that generate large changes from indefinitely small differences, as bifurcation does, must ultimately fail to characterize it accurately. We do not consider this argument final, but this is not the place to pursue the issue and it is anyway likely premature to do so at this juncture.

Realists need interpretable access to dynamics. They can simply choose to live within the limits, often quite severe limits, on this access, pragmatically constructing interpretable models whenever possible. But while this may be a valid policy for the setting,[7] the preceding considerations raise again the issue of what the criteria are for interpretable models. Are they simply the entities and relations that we humans find sufficiently convenient to use, say because of the way our perception and action has evolved? Or is there something more species-transcending (and so objective) behind their characterization? Suppose a Martian were to claim, as someday an ML robot might, that it was comfortable employing the hundreds-of-pages conception of gravity. Does it have an interpretable model? Where does it leave our notion of an interpretable model? It is not enough to say that it is what a theory quantifies over. This only shifts the question back to what constitutes an intelligible theory. Is it one that quantifies over entities intelligible to us?[8] These questions loom larger than this issue, reaching out to considerations of how diversity in cognitive capacities ought to bear upon accounts of objectivity, adequate translation, and so on. Yet they are also relevant here, since ML bids fair to being a domain of distinctive cognitive capacities that is destined to expand and deepen, and to intertwine itself with us in myriad ways.

7. *Conclusion*

The inherent value of an ontological description remains, for us, a compelling defence of realism. However, access to it is limited, and successful naked prediction challenges traditional conceptions of the scientific method and philosophy of science, whether realist or not. We expect these challenges to grow as machine learning presents an alternative to an ever-wider variety of tasks, resulting in a mode of scientific advance that is alien to our present philosophical conceptions. This will change the terms of the realist debate and many others.

---

[7] This is where one of us [CAH] left the discussion in a first essay into the domain (Hooker 2011) despite the informal efforts of the other [GJH].

[8] Compare the variety of conceptions of model interpretability as some form of intuitability that Lipton (2016) reveals.

Giles Hooker
Department of Biological Statistics and Computational Biology
Cornell University
Ithaca, NY 14850
Pmail: 1186 Comstock Hall, Cornell University, Ithaca, NY 14853
Email: gjh27@cornell.edu

Cliff. Hooker
Department of Philosophy
Newcastle University
Callaghan, NSW 2308
Australia
Pmail: as above
Email: Cliff.Hooker@newcastle.edu.au